\def\year{2020}\relax
\documentclass[letterpaper]{article} 
\usepackage{aaai20}  
\usepackage{times}  
\usepackage{helvet} 
\usepackage{courier}  
\usepackage[hyphens]{url}  
\usepackage{graphicx} 
\usepackage{graphicx}  

\usepackage{latexsym}
\usepackage{pbox}

\usepackage[utf8]{inputenc}

\usepackage{fixltx2e}
\usepackage{textcomp}
\usepackage{array}
\usepackage{amsmath}
\usepackage{amssymb}
\usepackage{multirow}
\usepackage{floatrow}
\usepackage{soul}
\usepackage{longtable}
\usepackage{listingsutf8}
\usepackage{empheq}
\usepackage{makecell}
\usepackage{tabularx}
\usepackage{etoolbox}
\usepackage{pgfplots}
\usepackage{booktabs}

\usepackage{amsmath}
\usepackage{algorithm}
\usepackage[noend]{algpseudocode}

\newcommand{\citet}[1]{\citeauthor{#1} \shortcite{#1}}
\newcommand{\citep}{\cite}

\title{Two Birds with One Stone: Investigating Invertible Neural Networks for Inverse Problems in Morphology}
\author{Gözde Gül Şahin, Iryna Gurevych\\ 
Research Training Group AIPHES and UKP Lab, \\ 
Department of Computer Science, Technische Universität Darmstadt\\
www.ukp.tu-darmstadt.de \\
www.aiphes.tu-darmstadt.de \\ 
}

\begin{document}

\maketitle

\begin{abstract}
    Most problems in natural language processing can be approximated as inverse problems such as analysis and generation at variety of levels from morphological \textit{(e.g., cat+Plural$\leftrightarrow$cats)} to semantic \textit{(e.g., (call + 1 2)$\leftrightarrow$``Calculate one plus two.'')}. 
    Although the tasks in both directions are closely related, general approach in the field has been to design separate models specific for each task.
    However, having one shared model for both tasks, would help the researchers exploit the common knowledge among these problems with reduced time and memory requirements.
    We investigate a specific class of neural networks, called Invertible Neural Networks (INNs)~\cite{ardizzone2018analyzing} that enable simultaneous optimization in both directions, hence allow addressing of inverse problems via a single model. 
    In this study, we investigate INNs on morphological problems casted as inverse problems. We apply INNs to various morphological tasks with varying ambiguity and show that they provide competitive performance in both directions. We show that they are able to recover the morphological input parameters, i.e., predicting the lemma (e.g., cat) or the morphological tags (e.g., Plural) when run in the reverse direction, without any significant performance drop in the forward direction, i.e., predicting the surface form (e.g., cats). 
\end{abstract}
\section{Introduction}

  Inverse problem is a general term used in natural sciences and mathematics to describe the process of recovering the hidden model parameters, $\mathbf{x}$, from a set of observations, $\mathbf{y}$. In general, the forward problem, i.e., generating observations/outputs from parameters, is well-defined; while the inverse problem is generally ill-posed, i.e., no (unique) solution exists. For instance, inferring seismic properties of the Earth’s interior from surface observations is a typical inverse problem in geophysics~\cite{snieder1999inverse}; since forward problem is well-defined and can be simulated by a forward model, recovering the seismic properties that lead to a specific surface value is ill-posed. Although inverse problems have been tackled in many fields such as imaging~\cite{bertero1998introduction,adler2019uncertainty}, astronomy~\cite{osborne2019radynversion,ardizzone2018analyzing} and geophysics~\cite{snieder1999inverse}, natural language processing (NLP) has not yet witnessed an explicit exploration, mostly due to the discrete nature of human language. However NLP contains many tasks that resemble inverse problems such as semantic parsing $\leftrightarrow$ text generation, morphological analysis $\leftrightarrow$ morphological generation, text-to-data (e.g., database records) $\leftrightarrow$ data-to-text generation and many more. Recently, the NLP field has replaced traditional discrete representations of text with dense and low-dimensional continuous ones, referred to as word/sentence vectors. This has enabled us to reformulate some of the classical morphological tasks as inverse problems within an existing framework that has been previously employed for such problems in other fields.  

  Inverse problems always exist together with their forward problem. Even so, traditional methods in NLP, optimize the inverse problems in an isolated fashion via a supervised loss for direct posterior probability learning, $p(\mathbf{x} \,|\, \mathbf{y})$, causing challenges for ill-posed problems, i.e., multiple possible $\mathbf{x}$ values in $\mathbf{y}\rightarrow\mathbf{x}$~\footnote{In case of a unique solution, we refer to it as a well-posed inverse problem, otherwise it is named as ill-posed following~\citet{kabanikhin2008definitions}}. In addition, the direct formulation ignores the connection to its forward problem, losing the potential to exploit the shared knowledge between those two. Furthermore, direct optimization of those problems requires two dedicated models to be trained separately, increasing the computational requirements. 
  
  Recently, a new class of neural networks, called Invertible Neural Networks (INNs), have been introduced by~\citet{ardizzone2018analyzing} following the research line of~\citet{dinh2014nice,DinhSB17,kingma2018glow}. INNs offer three key functionalities: (i) modeling inverse problems within a single network (shared parameters), (ii) having an invertible architecture that allows getting the inverse mapping $\mathbf{y}\rightarrow\mathbf{x}$ for free during prediction; and enables efficient bi-directional training \textit{(i.e., training the network at both ends)}; and (iii) addressing the ill-posed problems via an additional latent output variable $\mathbf{z}$ to have a one-to-one input and output mapping as $\mathbf{x}\leftrightarrow[\mathbf{y}, \mathbf{z}]$.
  \begin{figure*}[!ht]
	\centering
	\includegraphics[width=.95\columnwidth]{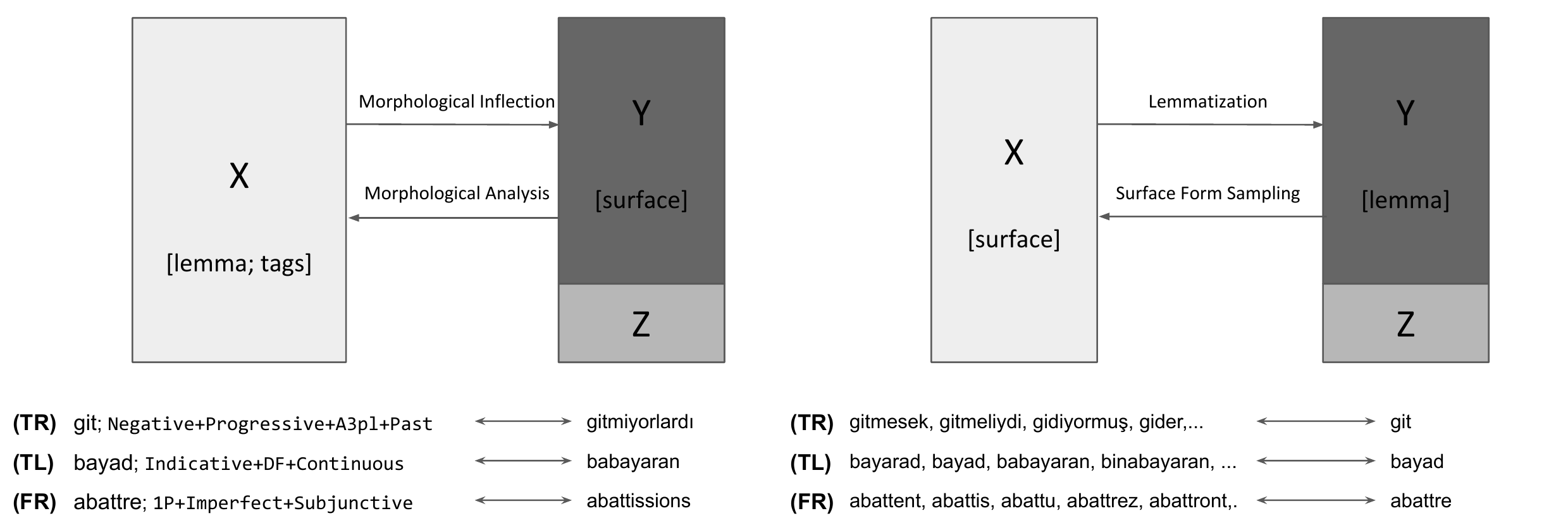}    
	\caption{Modeling of morphological tasks within the INN framework. \textbf{TR:} Turkish, \textbf{TL:} Tagalog, \textbf{FR:} French}
	\label{fig:inn_tasks}
  \end{figure*} 
  In this work, we model two well-known morphological tasks: \textit{morphological inflection} and \textit{lemmatization} together with their approximated inverses via modified invertible neural networks. \textit{Morphological inflection} is defined as inflecting a lemma with a set of inflectional morphemes to generate a surface form. Some examples of morphological inflection for Turkish, Tagalog and French are given in the left side of Fig.~\ref{fig:inn_tasks}. It is considered as a well-defined problem, since (almost always) only one surface form can be generated given the lemma and the tag combination. Its inverse problem is \textit{morphological analysis}, that aims to extract the lemma and the tags given the surface form. Unlike inflection, it may be ambiguous, i.e. there may be multiple possible analysis of a single surface form, depending on the linguistic properties of the language. Note that the examples in Fig.~\ref{fig:inn_tasks}-left have only one possible analysis. \textit{Lemmatization} is the task of finding the lemma of a given surface form. Although it is possible to have multiple lemmas per surface form, it is treated as a well-defined task in literature~\footnote{For instance, the Turkish word ``dolar'' can have the lemma ``dol'' (to fill), ``dola'' (to wrap) or ``dolar'' (dollar), which can only be determined when the surface form ``dolar'' is given within a sentence. However this task is considered different and named contextual lemmatization. Therefore such cases are not handled in scope of the lemmatization task.}. Its inverse problem is surface form generation, similar to morphological inflection, however without the guidance of morphological tags. We name it as surface form sampling, since one can sample a $z$ variable and generate surface forms given the lemma as shown in Fig.~\ref{fig:inn_tasks}-right. 

  In this work, we recognize, for the first time, that NLP contains problems similar to ``inverse problems'' that are found in many other fields such as imaging, geophysics, medical and astronomy. This recognition would provide new ways to approach traditional NLP problems by adapting the existing solutions developed for inverse problems in other fields. We experiment with several languages from diverse language families and morphological typologies and show that: 
  \begin{itemize}
    \item INNs have the ability to optimize for both problems simultaneously providing strong results for both, however generally 2\%-6\% less then a strong method designed specifically for the forward task under our experimental settings,
    \item Bi-directional training is \emph{crucial} to provide competitive scores for both sides of the network, e.g., adding reverse training to morphological inflection, boosts the lemma prediction performance (inverse problem), while causing only a slight drop in morphological inflection (forward process), 
    \item Introduction of additional categorical latent variables provide improvements in lemmatization for \textit{all languages}, even surpassing a strong model for some languages,
    \item INNs implicitly learn simple morphological tag distributions even without a dedicated loss function, however words needed direct supervision.  
  \end{itemize}
  We believe that this initial exploration of INNs for inverse problems in morphology would encourage research in that direction for more complex NLP tasks.

\section{Background: Invertible Neural Networks (INNs)} 
\label{sec:INN}
\begin{figure}[!ht]
  \centering
  \includegraphics[width=.65\columnwidth]{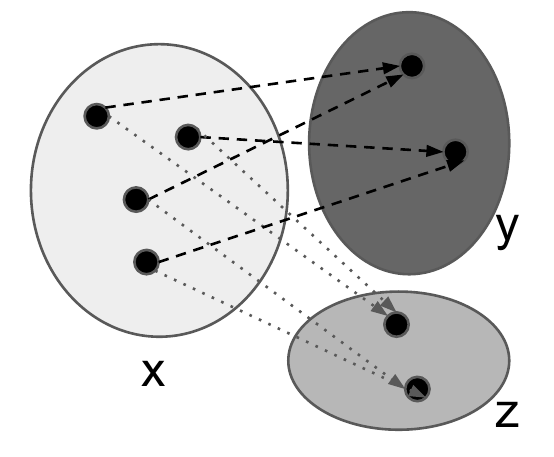}    
  \caption{Additional latent variable $\mathbf{z}$ for one-to-one mapping}
  \label{fig:inn_idea}
  \end{figure}

  INNs are originally proposed to address ill-posed inverse problems that have a well-defined forward process, i.e., a unique $x \rightarrow y$, whereas the inverse problem is ambiguous, i.e., multiple $y \rightarrow x$ mappings. It is addressed by introducing an additional latent variable, 
  as illustrated in Fig.~\ref{fig:inn_idea}, that turns $\mathbf{x} \leftrightarrow [\mathbf{y}, \mathbf{z}]$ into a bijective mapping. In other words, it enables the reparametrization of $p(\mathbf{x} \,|\, \mathbf{y})$ into a deterministic function $\mathbf{x} = f(\mathbf{y}, \mathbf{z})$. Therefore we can define an inverse function $x = f^{-1} (\mathbf{y}, \mathbf{z}) = g(\mathbf{y}, \mathbf{z})$ to estimate the full posterior distribution, $p(\mathbf{x} \,|\, \mathbf{y})$.
  Recently proposed neural network components that are invertible such as coupling layers can now be employed as described in more details later in this section. Such networks are unique since they can be run backwards to get the inverse $\mathbf{y} \rightarrow \mathbf{x}$ without any additional cost at prediction time. 
  Furthmermore, INNs offer efficient training procedure that can optimize both ends of the network simultaneously, which is referred to as bi-directional training. 

  \subsection{Invertible Component}
  \label{ssec:coupling}
    INNs consist of layers of ``reversible block''s shown with Fig.~\ref{fig:inn_arch} that contain complemantary affine coupling layers proposed by~\citet{DinhSB17}. First, the input vector $u$ is split into two halves $\mathbf{u}_1$ and $\mathbf{u}_2$ (in a fixed way
    ), which are then transformed by the learned affine functions $s_i$ and $t_i$, where $i \in \{ 1,2 \}$ using the following equations: 
    \begin{align*} \mathbf{v}_1 &= \mathbf{u}_1 \odot \exp\!\big(s_2(\mathbf{u}_2)\big) + t_2(\mathbf{u}_2) \\ \mathbf{v}_2 &= \mathbf{u}_2 \odot \exp\!\big(s_1(\mathbf{v}_1)\big) + t_1(\mathbf{v}_1), \end{align*}
    where $\odot$ denotes element-wise multiplication. The output $\mathbf{v}$ is then simply calculated by concatenating $\mathbf{v}_1;\mathbf{v}_2$. The input $\mathbf{u}$ can then be recovered with:
    \begin{align*} \mathbf{u}_2 &= (\mathbf{v}_2 - t_1(\mathbf{v}_1)) \odot \exp\!\big(-s_1(\mathbf{v}_1)\big) \\ \mathbf{u}_1 &= (\mathbf{v}_1 - t_2(\mathbf{u}_2)) \odot \exp\!\big(-s_2(\mathbf{u}_2)\big). \end{align*}
    $s_i$ and $t_i$ need not be invertible and generally represented by multi-layered feed-forward neural networks with nonlinear activations, which can then be trained by standard backpropogation algorithm. Similar to previous methods, in order to vary the splits $[\mathbf{u}_1, \mathbf{u}_2]$ among different layers, we employ a permutation layer between the reversible blocks which shuffle the elements in a randomized but a fixed way. 

  \subsection{Bi-directional Training}
  \label{ssec:training}
\begin{figure}
      \centering
      \includegraphics[width=.70\columnwidth]{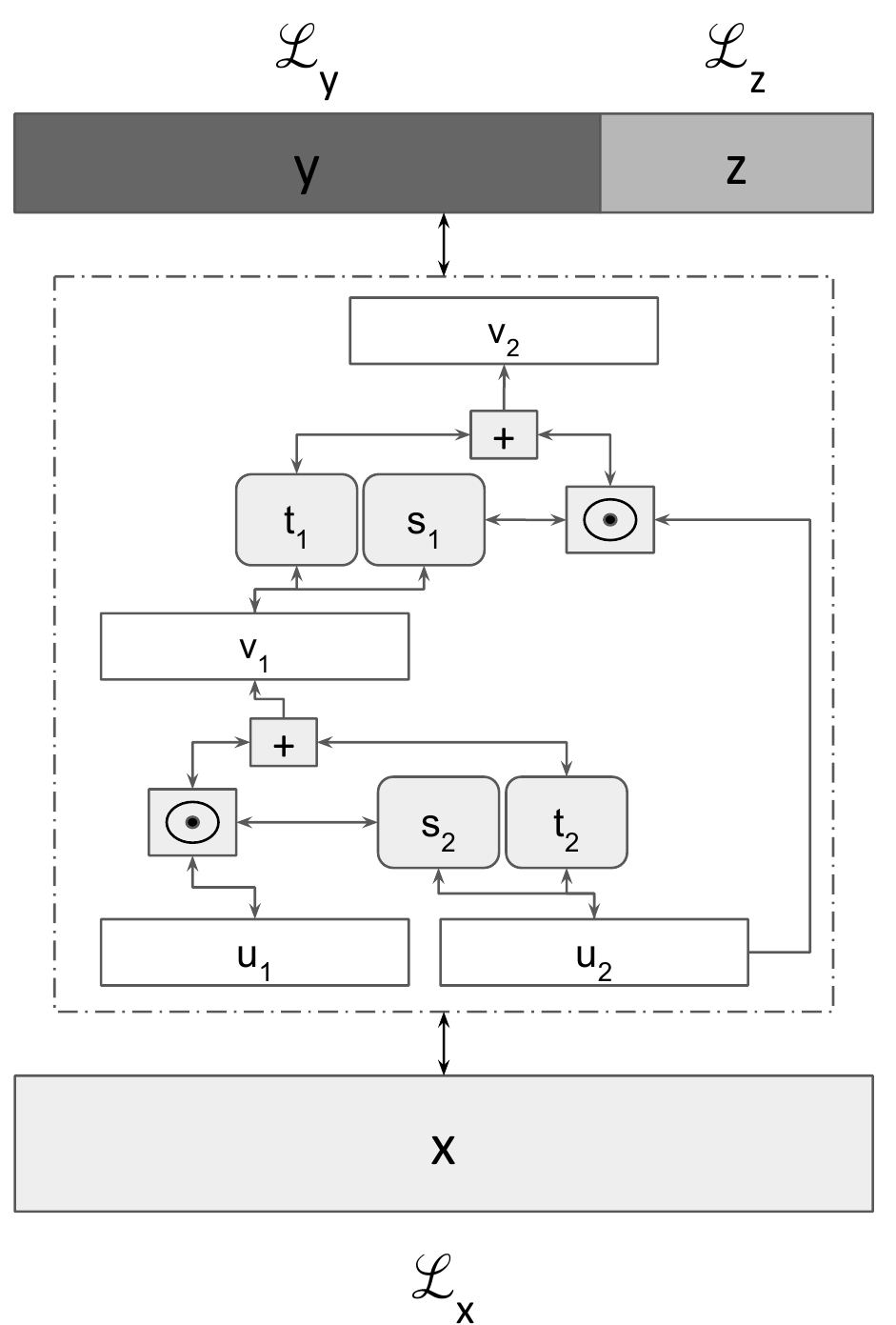}    
      \caption{Demonstration of one layer INN. Revertible block is shown inside the dashed box. The operations (e.g., $+, \odot$) are only shown for the forward direction for convenience. They are inverted to subtraction and element-wise division in reverse direction. $\mathcal{L}$ denotes loss function.}
    \label{fig:inn_arch}
    \end{figure}

    As discussed by~\citet{GroverDE18}, networks that are invertible offer a unique oppurtunity to optimize for both the input and output domains \textit{simultaneously}, i.e., apply the loss functions $\mathcal{L}_x$, $\mathcal{L}_y$ and $\mathcal{L}_z$ given in Fig.~\ref{fig:inn_arch} for the forward and the inverse passes at the same time. 
    This is achieved by performing forward and inverse iterations in an alternating fashion, and then accumulating the gradients to update the network parameters. This training procedure has been shown to be highly beneficial in auto-encoders~\cite{teng2019invertible} and our own experiments. 

\section{Method}
\label{sec:method}
    Different applications of INNs~\cite{osborne2019radynversion,ardizzone2018analyzing,adler2019uncertainty} have commonly used continuous latent variables with Gaussian priors as $\mathbf{z} \sim \mathcal{N}(\mu,\,\sigma^{2})$. 
    Unlike in previously addressed ill-posed inverse problems, most NLP tasks contain categorical and discrete values. For instance in lemmatization the $\mathbf{z}$ variable is expected to capture the morphological tags which are categorical by definition. Therefore, we have employed categorical latent variables~\cite{DBLP:conf/iclr/JangGP17} that is based on the idea of approximating a categorical distribution via a differentiable distribution called \textit{Gumbel-Softmax (GS)}. This distribution can be smoothly annealed into a categorical distribution via a temperature parameter, $\tau$. As $\tau \rightarrow 0$, the GS distributions are identical to samples from a categorical distribution (one-hot), where as $\tau \rightarrow \infty$ GS samples become uniform. 

    Then the method can be formally defined as: Given the input vector $\mathbf{x} \in \mathbb{R}^n$, the output vector $\mathbf{y} \in \mathbb{R}^m$, we assume that the forward process $\mathbf{y}=\mathbf{s}(\mathbf{x})$, \textit{i.e., morphological inflection and lemmatization}, is well-defined by an arbitrary transformation function $\mathbf{s}$. Our goal is to approximate the posterior $p(\mathbf{x} \,|\, \mathbf{y})$ by a tractable distribution $q(\mathbf{x} \,|\, \mathbf{y})$. This is then reparameterized by a deterministic function $g$, represented by the neural network with parameters $\theta$. Categorical output latent variable $\mathbf{z} \in \mathbb{R}^{d \times cat}$ is drawn from a \textit{Gumbel-Softmax (GS)} distribution, where $d$ is latent variable dimension and $cat$ is the number of categories for each dimension. Then $g$ is defined as:
    \begin{align*} \mathbf{x} = g (\mathbf{y},\mathbf{z};\theta) \quad \textrm{where} \quad  \mathbf{z} \approx p(z) = GS(z;\tau) \end{align*} 
    We learn the inverse model $g (\mathbf{y},\mathbf{z};\theta)$ jointly with the forward model $f (\mathbf{x}; \theta)$ that approximates the $s(\mathbf{x})$ function:  
    \begin{align*} [\mathbf{y},\mathbf{z}] = f (\mathbf{x};\theta) = [f_y(\mathbf{x};\theta), f_z(\mathbf{x};\theta)] = g^{-1}(\mathbf{x};\theta) \\ \textrm{and} \quad f_y(\mathbf{x};\theta) \approx s(\mathbf{x}) \end{align*} 
    It should be noted that both functions $f$ and $g$ share the same parameters $\theta$, therefore can be implemented by a \textbf{single} model. The invertibility property, $f = g^{-1}$ is ensured by the architecture that consists of invertible affine coupling layers defined in previous background section. One of the restrictions of using an invertible architecture is the equality constraint on the dimensions of the input $\mathbf{x}$ and the output $[\mathbf{y},\mathbf{z}]$. To satisfy this constraint, we pad the lower-dimensional side (i.e., input or output) with zeros~\footnote{One could implement an additional loss to keep padding values closer to zero throughout the training. However we have not seen any significant change in performances when implemented.}. Finally the posterior distribution is calculated as the following as shown in~\cite{ardizzone2018analyzing}: 
    \begin{align*} q(\mathbf{x}=g(\mathbf{y},\mathbf{z};\theta)|\mathbf{y}) = p(\mathbf{z}) |J_x|^{-1}, \textrm{where} \\J_x = det \left( \frac{\partial {g(\mathbf{y},\mathbf{z}|\theta)}}{\partial {[\mathbf{y},\mathbf{z}]}}\bigg|_{\mathbf{y},f_z(\mathbf{x})} \right) \end{align*}
    where $J_x$ is the Jacobian determinant and can be easily calculated due to its triangular structure. We refer the reader to the original study \cite{ardizzone2018analyzing} for details of the above derivations. 

    For all morphological tasks, the words (lemmas and surface forms) are divided into smaller units, i.e., subword units, via the Byte-Pair-Encoding (BPE) algorithm~\cite{SennrichHB16a}. We then used the 100 dimensional pretrained multilingual embeddings released by~\citet{heinzerling2018bpemb} that provide vector representations for the BPE subword units. Finally each word is represented as sum of its subword unit vectors that we denote as $\vec{lemma} \in \mathbb{R}^{100}$ and $\vec{surface} \in \mathbb{R}^{100}$. For \textit{morphological inflection}, we define the vector $\vec{t} \in [0,1]^N$ to represent the morphological tagset, where $N$ is the number of distinct morphological tags in the training set. Then $t_i$ is set to $1$, if the feature $i$ is among the tagset of the surface form. 


    \paragraph{Variables:} For \textit{morphological inflection}, as given in Fig.~\ref{fig:inn_tasks}-left, $\mathbf{x}$ is defined as $\mathbf{x} = [\vec{lemma};\vec{t}]$ where $;$ denotes concatentation operation and $\mathbf{y}$ is simply equal to $\vec{surface}$. According to Fig.~\ref{fig:inn_tasks}-right, $\mathbf{x}$ and $\mathbf{y}$ are equal to $\vec{surface}$ and $\vec{lemma}$ accordingly for \textit{lemmatization}.

	\paragraph{Loss Functions:} We use a supervised loss, implemented via cosine distance function, to minimize the error between the network prediction $\vec{surface'}$ and the gold value $\vec{surface}$ defined as: \begin{align*} \mathcal{L}_{surface} = 1 - \frac{\vec{surface} \cdot \vec{surface'}}{\|\vec{surface}\| \|\vec{surface'}\|}\end{align*} for both tasks. The loss between the predicted $\vec{lemma'}$ and the gold $\vec{lemma}$ is again calculated via cosine distance. For morphological inflection where $\vec{t'}$ is the predicted and $\vec{t}$ is the gold tag vector, we minimize the error using binary cross entropy loss defined as below: \begin{align*} \mathcal{L}_t = - \frac{1}{N} \sum_{i=0}^{N} (t_i.log(t'_i) + (1-t_i) log(1-t'_i)) \end{align*} $\mathcal{L}_x$ in Fig.~\ref{fig:inn_arch} is defined as $\mathcal{L}_{lemma} + \mathcal{L}_t$ for \textit{morphological inflection}, where it is equal to $\mathcal{L}_{surface}$ for \textit{lemmatization} task. Similarly $\mathcal{L}_y$ is defined as $\mathcal{L}_{surface}$ and $\mathcal{L}_{lemma}$ respectively for \textit{morphological inflection} and \textit{lemmatization}. Finally, we use KL divergence loss for $\mathbf{z}$. The final losses are calculated as following:
	\begin{align*} \mathcal{L}_{inflection} = \alpha_x (\mathcal{L}_{lemma}+\mathcal{L}_{t})+ \alpha_y \mathcal{L}_{y} + \alpha_z \mathcal{L}_{z} \\ \mathcal{L}_{lemmatization} = \alpha_x \mathcal{L}_{x}+ \alpha_y \mathcal{L}_{y} + \alpha_z \mathcal{L}_{z} \end{align*} The relative weights of the losses denoted with $\alpha$ are adjusted as hyperparameters. 
   
    \paragraph{Training Procedure:} Bi-directional training is performed iteratively to calculate forward and backward losses. The pseudocode for training \textit{morphological inflection} task is given in Algorithm~\ref{fig:pseudo}. 
    \begin{algorithm}
	\caption{Training Procedure}
		\begin{algorithmic}[1]
		\Procedure{TrainInflection}{}
		\For{\textbf{each} \textit{lemma}, \textit{tagset}, \textit{surface} in train\_split}
		\State $\mathbf{x} \gets [\vec{lemma};\vec{t}]$
		\State $\mathbf{y} \gets \vec{surface}$
		\State $\mathbf{y'}, \mathbf{z'} \gets INN(\mathbf{x})$
		\State $\mathbf{z} \gets GumbelSoftmax (\mathbf{z'})$
		\State $\mathbf{x'} \gets INN(\mathbf{y,z}, \textbf{reverse}=\texttt{True})$
		\State $\mathcal{L}_{total} \gets \alpha_x \mathcal{L}_{x}+ \alpha_y \mathcal{L}_{y} + \alpha_z \mathcal{L}_{z} $	
        \State $\mathcal{L}_{total}.backward()$	
        \EndFor
		\EndProcedure
		\end{algorithmic}
	\label{fig:pseudo}
	\end{algorithm}

	\paragraph{Testing Procedure:} In both models, we choose the word with the highest cosine similarity to the predicted vector during testing. This nearest neighbor search is very efficiently implemented in the gensim framework~\cite{rehurek_lrec}. To predict morphological tagset, we use sigmoid activation function on $\vec{t'}$ and choose the features with activations above $0.5$.

\section{Experiments}
  We first describe the dataset used in the paper, then detail our experimental design, training settings and evaluation measures.  

  \subsection{Dataset}
  \label{ssec:datasets}
   
    \textbf{Wicentowski} is the most commonly used dataset for lemmatization, which is one of the most interesting ill-posed inverse problems investigated in this paper. Previous studies~\cite{Dreyer2011,rastogi2016weighting} use a subset of the dataset created by \citet{Wicentowski:2003:MLM:936459}~\footnote{We thank the author for sharing the dataset with us.}. However, since previous works evaluate on the lemmatization task only, this subset does not contain any morphological tags but only the unique lemma-surface pairs. Therefore we have used the original dataset that additionally contains morphological tags. In this dataset, each language has its own set of morphological tags. Furthermore, they are arbitrarily written (e.g., 1P and Past instead of person=``1P'' and tense=``Past'') without providing the morphological category like \textit{person} and \textit{tense}. The statistics for the dataset is given in Table~\ref{tab:data-stat}. 
  \begin{table}[!ht]
  \centering
  \resizebox{.95\columnwidth}{!}{
    \begin{tabular}{ccccc}
      \toprule
      \multirow{2}{*}{Dataset} & \multicolumn{3}{c}{Number of tokens} & \#Tag\\
      \cmidrule{2-4}
       & train & dev & test \\
      \midrule 
            \textit{FR}  & 81K/14K & 10K/2K & 10K/2K & 20 \\
            \textit{FIN} & 70K/46K & 9K/6K & 9K/6K & 35 \\
            \textit{RO}  & 41K/5K & 5K/1K & 5K/1K & 23 \\
            \textit{TR}  & 6.3K/1.6K & 782/201 & 783/202 & 27 \\
            \textit{TL}  & 1.4K/1.3K & 184/165 & 184/165 & 25 \\
            \textit{GA}  & 864/550 & 108/69 & 108/69 & 17 \\
      \bottomrule
    \end{tabular}
    }
    \caption{Dataset Statistics. First half: Celex data, Second half: Wicentowski dara. \textit{FR}: French, \textit{FIN}: Finnish, \textit{RO}: Romanian, \textit{TR}: Turkish, \textit{GA}: Irish. Numbers after ``/'' show the number of unique lemma-surface pairs. \#Tag: Number of unique morphological tags in the training data.}
    \label{tab:data-stat}
    \end{table}
   \begin{table}
      \centering
      \resizebox{.95\columnwidth}{!}
      {
          \begin{tabular}{lllrrr}
            \toprule
             &  & \bf{Model} & \bf{L (EM\%)} & \bf{Tag (F1\%)} & \bf{S (EM\%)} \\
            \toprule
            \multirow{9}{*}{\rotatebox[origin=c]{90}{Finnish}} & \multirow{5}{*}{\rotatebox[origin=c]{90}{\textsc{Lem}}} & baseline & 94.0 & - & - \\
            & & \citep{AharoniG17} & \textbf{99.6} & - & - \\
            \cmidrule{3-6}
            & & INN ($\mathcal{L}_y$+$\mathcal{L}_x$) & 97.68 & - & \\
            & & INN (+$\mathcal{L}_z$, $dim_z$=2, $dim_{cat}$=3) & 97.54 & - & \\
            & & INN (+$\mathcal{L}_z$, $dim_z$=6, $dim_{cat}$=4) & \textit{98.18} & - & \\
            \cmidrule{2-6}
            & \multirow{4}{*}{\rotatebox[origin=c]{90}{\textsc{Inf}}} & baseline & - & - & 85.15 \\
            \cmidrule{3-6}
            & & INN ($\mathcal{L}_y$) & 0.01 & 11.84 & \textbf{94.07} \\
            & & INN ($\mathcal{L}_y$+$\mathcal{L}_x$) & \textbf{95.56} & 12.42 & 92.23 \\
            & & INN ($\mathcal{L}_y$+$\mathcal{L}_x$+$\mathcal{L}_t$) & 93.33 & \textbf{49.99} & 91.26 \\
            \midrule
            \multirow{9}{*}{\rotatebox[origin=c]{90}{French}} & \multirow{5}{*}{\rotatebox[origin=c]{90}{\textsc{Lem}}} & baseline & 95.91 & - & - \\
            & & \citep{AharoniG17} & \textbf{98.24} & - & - \\
            \cmidrule{3-6}
            & & INN ($\mathcal{L}_y$+$\mathcal{L}_x$) & 95.91 & - & \\
            & & INN (+$\mathcal{L}_z$, $dim_z$=2, $dim_{cat}$=3) & 96.03 & - & \\
            & & INN (+$\mathcal{L}_z$, $dim_z$=6, $dim_{cat}$=4) & \textit{96.14} & - & \\
            \cmidrule{2-6}
            & \multirow{4}{*}{\rotatebox[origin=c]{90}{\textsc{Inf}}} & baseline & - & - & 88.94 \\
            \cmidrule{3-6}
            & & INN ($\mathcal{L}_y$) & 0.0 & 23.5 & \textbf{98.18} \\
            & & INN ($\mathcal{L}_y$+$\mathcal{L}_x$) & \textbf{98.56} & 13.32 & 97.55 \\
            & & INN ($\mathcal{L}_y$+$\mathcal{L}_x$+$\mathcal{L}_t$) & 98.27 & \textbf{26.01} & 97.49 \\
            \midrule
            \multirow{9}{*}{\rotatebox[origin=c]{90}{Irish}} & \multirow{5}{*}{\rotatebox[origin=c]{90}{\textsc{Lem}}} & baseline & \textit{95.35} & - & - \\
            & & \citep{AharoniG17} & 94.2 & - & - \\
            \cmidrule{3-6}
            & & INN ($\mathcal{L}_y$+$\mathcal{L}_x$) & 94.2 & - & \\
            & & INN (+$\mathcal{L}_z$, $dim_z$=2, $dim_{cat}$=3) & 94.2 & - & \\
            & & INN (+$\mathcal{L}_z$, $dim_z$=6, $dim_{cat}$=4) & \textbf{95.65} & - & \\
            \cmidrule{2-6}
            & \multirow{4}{*}{\rotatebox[origin=c]{90}{\textsc{Inf}}} & baseline & - & - & 61.11 \\
            \cmidrule{3-6}
            & & INN ($\mathcal{L}_y$) & 0.0 & 25.98 & 76.85 \\
            & & INN ($\mathcal{L}_y$+$\mathcal{L}_x$) & 97.22 & 20.55 & \textbf{79.63} \\
            & & INN ($\mathcal{L}_y$+$\mathcal{L}_x$+$\mathcal{L}_t$) & \textbf{100} & \textbf{67.33} & 77.78 \\
            \midrule
            \multirow{9}{*}{\rotatebox[origin=c]{90}{Romanian}} & \multirow{5}{*}{\rotatebox[origin=c]{90}{\textsc{Lem}}} & baseline & 82.39 & - & - \\
            & & \citep{AharoniG17} & \textbf{92.89} & - & - \\
            \cmidrule{3-6}
            & & INN ($\mathcal{L}_y$+$\mathcal{L}_x$) & 89.5 & - & \\
            & & INN (+$\mathcal{L}_z$, $dim_z$=2, $dim_{cat}$=3) & 88.85 & - & \\
            & & INN (+$\mathcal{L}_z$, $dim_z$=6, $dim_{cat}$=4) & \textit{90.95} & - & \\
            \cmidrule{2-6}
            & \multirow{4}{*}{\rotatebox[origin=c]{90}{\textsc{Inf}}} & baseline & - & - & 89.45 \\
            \cmidrule{3-6}
            & & INN ($\mathcal{L}_y$) & 0.0 & \textbf{16.5} & \textbf{97.87} \\
            & & INN ($\mathcal{L}_y$+$\mathcal{L}_x$) & \textbf{99.26} & 14.24 & 97.67 \\
            & & INN ($\mathcal{L}_y$+$\mathcal{L}_x$+$\mathcal{L}_t$) & 98.88 & 0.02 & 97.34 \\
            \midrule
            \multirow{9}{*}{\rotatebox[origin=c]{90}{Tagalog}} & \multirow{5}{*}{\rotatebox[origin=c]{90}{\textsc{Lem}}} & baseline & 88.48 & - & - \\
            & & \citep{AharoniG17} & \textbf{92.72} & - & - \\
            \cmidrule{3-6}
            & & INN ($\mathcal{L}_y$+$\mathcal{L}_x$) & 88.06 & - & \\
            & & INN (+$\mathcal{L}_z$, $dim_z$=2, $dim_{cat}$=3) & 90.03 & - & \\
            & & INN (+$\mathcal{L}_z$, $dim_z$=6, $dim_{cat}$=4) & \textit{91.52} & - & \\
            \cmidrule{2-6}
            & \multirow{4}{*}{\rotatebox[origin=c]{90}{\textsc{Inf}}} & baseline & - & - & 30.05 \\
            \cmidrule{3-6}
            & & INN ($\mathcal{L}_y$) & 0.0 & 15.24 & \textbf{37.7} \\
            & & INN ($\mathcal{L}_y$+$\mathcal{L}_x$) & \textbf{85.41} & 10.13 & 33.33 \\
            & & INN ($\mathcal{L}_y$+$\mathcal{L}_x$+$\mathcal{L}_t$) & 69.73 & \textbf{43.14} & 33.88 \\
            \midrule
            \multirow{9}{*}{\rotatebox[origin=c]{90}{Turkish}} & \multirow{5}{*}{\rotatebox[origin=c]{90}{\textsc{Lem}}} & baseline & 95.28 & - & - \\
            & & \citep{AharoniG17} & 96.53 & - & - \\
            \cmidrule{3-6}
            & & INN ($\mathcal{L}_y$+$\mathcal{L}_x$) & 98.51 & - & \\
            & & INN (+$\mathcal{L}_z$, $dim_z$=2, $dim_{cat}$=3) & 99.01 & - & \\
            & & INN (+$\mathcal{L}_z$, $dim_z$=6, $dim_{cat}$=4) & \textbf{99.54} & - & \\
            \cmidrule{2-6}
            & \multirow{4}{*}{\rotatebox[origin=c]{90}{\textsc{Inf}}} & baseline & - & - & 88.89 \\
            \cmidrule{3-6}
            & & INN ($\mathcal{L}_y$) & 0.0 & 36.29 & \textbf{97.57} \\
            & & INN ($\mathcal{L}_y$+$\mathcal{L}_x$) & 99.62 & 24.37 & 97.45 \\
            & & INN ($\mathcal{L}_y$+$\mathcal{L}_x$+$\mathcal{L}_t$) & \textbf{99.74} & \textbf{77.34} & 97.45 \\
            \bottomrule
          \end{tabular}
          }
    \caption{Lemmatization and inflection results for different languages and experimental settings. \textsc{Lem}: Lemmatization, \textsc{Inf}: Inflection. L (EM\%): Lemma exact match score, S (EM\%): Surface exact match score. Best scores are given in \textbf{bold} for each score. Second best scores for lemmatization is shown in \textit{italics}. EM: Exact Match, $dim_z$, $dim_{cat}$: dimensions of $\mathbf{z}$ variable and number of categories for each.}
    \label{tab:wicentowski_res}
    \end{table}
  
  \subsection{Experimental Design}
  \label{ssec:exp_des}


    
    We perform morphological experiments with the proposed framework on 6 different languages from diverse language families and morphology types, namely Turkish (Turkic, agglutinative), Romanian (Romance, fusional), Finnish (Uralic, agglutinative), French (Romance, fusional), Irish (Celtic, fusional) and Tagalog (Austronesian, agglutinative). Agglutinative and fusional languages have different morphological properties. Typical properties of agglutinative languages include (1) the ability to generate/derive words by attaching morphemes like beads on a string; and (2) associating each morpheme with one certain semantic unit only, e.g., using the Turkish ``lar'' morpheme only for plurality. This leads to having a high ratio of out-of-vocabulary words, however the meaning of words are generally predictable. Unlike agglutinative languages, fusional languages typically have smaller ratio of morphemes per word, i.e., not as productive as agglutinative languages. However one morpheme is generally associated with multiple meaning units, e.g., tense and person information conveyed with one morpheme. This generally results in lower out-of-vocabulary ratios with more ambiguous words, especially when the context is unknown.    

  \subsection{Training and Evaluation}
  \label{ssec:train_eval}

    For all models, we have used 100-dim vectors extracted from pretrained Byte-Pair-Encoding (BPE) model with vocabulary size of 1K, provided by~\citet{heinzerling2018bpemb}. We have initialized the weight parameters orthogonally. For all INN models, we have used 3 invertible blocks, with 3 fully connected linear layers with hidden dimension of 128 (with ReLU activations in the intermediate layers) as affine coefficient function. We used gradient clipping and early stopping to prevent overfitting. Models are optimized with Adam optimizer with the initial learning rate of 0.001, decreased by 0.3 if scores on development set do not improve for 5 epochs. Unless otherwise stated, we used the weights $\alpha_x=20$, $\alpha_t=10$, $\alpha_y=80$ and $\alpha_z=1$ respectively for $\mathcal{L}_x$, $\mathcal{L}_t$, $\mathcal{L}_y$ and $\mathcal{L}_z$. These weights are intuitively chosen following the previous studies that has employed INNs. We have trained all models for 30 epochs, and did not perform a comprehensive hyperparameter search since the goal of this study is not delivering state-of-the-art results, rather providing and investigate a different perspective.

    For the baseline models, we used 3 fully connected linear layers with hidden dimension of 128, with ReLU activations in intermediate layers. We kept the other settings the same as the INNs (except from zero-padding, since it is not necessary for feed forward networks). Our method is not directly comparable to previous methods for three main reasons: (a) none of the previous morphological inflection or works report two (three when tag scores are included) scores with the same model, (b) none of the previous lemmatization studies offer sampling surface forms with the given lemma, (c) most of the previous works are able to generate unseen words, while we treat the generation as ``a selection from a large vocabulary'' problem that simplifies the problem space. Nevertheless, we have chosen one of the recent state-of-the-art models by~\citep{AharoniG17} as a reference for comparison, to provide an insight about the proposed model's performance. For evaluation we use the percentage of exact match score for lemma and surface form predictions, and F1 score for morphological tag predictions due to having multiple labels per form.

    Since a nearest neighbor search is performed between the network output and the existing word embedding space using cosine similarity, words that have not been encountered in the training data, would not be predicted. In order to address this, we have extended the vector space with 500K most common words, which can be considered quite many, along with the words encountered in the test data during prediction time. 

    As discussed previously, \textit{morphological analysis}, i.e., inverse of the \textit{morphological inflection} task, may be ill-posed depending on the linguistic properties of the language. For instance, agglutinative languages have one-to-one morpheme-to-tag mapping, while for fusional languages one morpheme may stand for multiple tags. This means that, \textit{morphological analysis} of fusional languages is ill-defined while (mostly) the opposite is true for agglutinative ones. In case of ill-posedness, continuous sampling from $\mathbf{z}$ is necessary to find distributions of the morphological tags, which are hard to score. To simplify scoring of predicted tags, we ignore $\mathbf{z}$ and only consider the most likely tagset in a current setting. Since the ambiguity greatly varies with the language families, we focus on the tag scores of languages with less ambiguous input.
  

\section{Results and Analysis}  
    We present the results of lemmatization and the morphological inflection tasks for 6 languages in Table~\ref{tab:wicentowski_res}. First, we observe that all models in all experimental settings outperform the baseline.     
    
    \paragraph{Lemmatization:} Inroducing latent variables help increasing the performance of lemmatization \textit{in all languages}, surpassing the \cite{AharoniG17} by a small margin for Turkish and Irish, and providing similar results for other languages. More specifically, relative performance to \cite{AharoniG17} ranges between [-2\%, +3\%]. In addition, we observe that larger the $z$ dimension gets, better the scores become in all languages. It may be due to simply providing the network with a larger representation space, allowing for a more flexible learning process. We have then used the trained model to sample surface forms for the given lemma. We have observed that, the model almost always generates a valid surface form when run backwards. However we have also noticed the diversity of the generated surface forms being quite low. This may be due to the statistical properties of the dataset, i.e., observing one dominant morphological tag combination throughout the trainig set. Second reason, is the problem similar to the one described in \citet{BowmanVVDJB16}. This study uses Variational Autoencoders (VAE) to generate diverse sentences via the help of latent variable, $z$, that aims to encode useful global information hence enable generation of diverse sentences. However, they observe \textit{KL-divergence} loss of zero that causes the model to ignore $z$. 
    
    \paragraph{Morphological Inflection:} One of the most important findings for this task is the evidence of network's ability to provide remarkably strong results for both directions when $\mathcal{L}_y$+$\mathcal{L}_x$ are used. This suggests that, optimizing the same network parameters ($\theta$) with loss functions of dual problems, is feasible and necessary for dual strong results. Although majority of the time, adding $\mathcal{L}_t$, slightly decreased the lemma recovery scores, in Irish and Turkish, we observe improvement on both scores, which may be due to relatively small number of training data. We see mixed results for tag F1 scores, due to varying morphological ambiguity in languages. For agglutinative languages, Finnish, Tagalog and Turkish, where each morpheme is associated with a tag, the ambiguity is lower, therefore F1 scores are higher; in contrast to fusional languages, Romanian and French. Interestingly, even when the network is only optimized for $\mathcal{L}_y$, INNs started to implicitly learn the morphological tags \emph{for all languages}, demonstrated by the tag F1 scores ranging between $11.84$-$36.29$. However no improvement had been observed for the lemma, suggesting that the distribution of the lemma is too complicated to learn implicitly, hence an explicit supervision is necessary. Finally, for all agglutinative languages, lemma exact match scores in \textit{lemmatization} task are better than or very similar to the scores in the inverse task of \textit{morphological analysis}; which can be explained by one-to-one morpheme to tag mapping, that is implicitly learned without any guidance.  
    
   
\section{Related Work}
  \paragraph{Morphological Tasks}
  Most morphological problems dealt in this paper are also known as string-to-string transduction problems. \citet{DreyerSE08} introduce a general modeling framework based on weighted finite state transducers (WFST) that employ n-gram features and latent variables. They perform experiments on morphological form generation and lemmatization and show that incorporating latent variables improves the results dramatically. Although they use the same framework for both problems, the models are trained separately for each problem. Furthermore, the framework learns the best alignment between the lemma and the inflected form, \emph{separately} for each task, which may not be known beforehand in a realistic scenario. ~\citet{SchnoberEDG16} compare more recent encoder-decoder architectures such as seq2seq with hard monotonic attention~\cite{AharoniG17} with traditional transduction techniques based on conditional random fields and WFSTs on classical string transduction tasks including lemmatization. They find that although traditional models have similar performance to neural models in most cases, their performances fall behind of the neural models for  lemmatization task. This suggests that the lemmatization can be considered as a more complex/ambiguous problem compared to other tasks. ~\citet{RibeiroNCC18} introduce a method to reduce the transduction to sequence labeling problem, and show that although neural methods perform on par or worse (e.g., on Finnish OCR) in most cases, it is the opposite for the morphological inflection task. 

  \paragraph{Invertibility in NLP}
  \citet{He_emnlp_18} employ invertible transformations (coupling layers that are very similar to the ones used in this study), to perform unsupervised learning of syntactic structure. They demonstrate the efficiency of the invertibility property on POS tagging and dependency parsing. Recently, \citet{ziegler2019latent} proposed a flow-based model for discrete sequences such as text, and show that their proposed autoregressive model performs on par with traditional sequential models on character-level language modeling task. 
 
  \paragraph{Joint models}
  A few decades ago, the researchers have attempted to design a unified framework, i.e., a reversible, single grammar, for parsing and generation~\cite{DBLP:conf/coling/Shieber88,wintner2000amalia}. However, the grammar development has been replaced by advanced statistical tools; therefore those works have not been explored recently. Another set of models that are conceptually similar to ours, perform paired training~\cite{DBLP:conf/acl/KonstasIYCZ17,DBLP:conf/icml/HuYLSX17,CaoZLLY19}, however still optimize separate models and generally have a more complicated training procedure. 

\section{Conclusions}  
    We have proposed modeling several inverse problems in morphology together with their dual problem, such as morphological analysis $\leftrightarrow$ inflection, with recently proposed invertible neural networks that uses a single model for both problems, offer efficient bi-directional training with the help of invertibility layers and provide free inverse mapping. We showed that they are capable of simultaneous optimization of such dual problems providing strong results on both; and lemmatization benefits from additional categorical latent variables. We demonstrated that simple distributions are implicitly learned while complex, multimodal distributions needed supervision. We hope that these initial encouraging results for inverse problems in morphology would inspire the researchers to explore other inverse problems of NLP.
\section{Acknowledgments}
This work has been supported by the German Research Foundation through the German-Israeli Project Cooperation (DIP, grant DA 1600/1-1 and grant GU 798/17-1). We would like to thank anonymous reviewers, Jonas Pfeiffer, Mohsen Mesgar, Ilia Kuznetsov and Yevgeniy Puzikov for their constructive feedback. We are grateful to Lynton Ardizzone, Jakob Kruse and Ullrich Köthe for the initial discussions and kindly sharing the INN implementation before the official release with us.  

\bibliography{emnlp-ijcnlp-2019}
\bibliographystyle{aaai}

\end{document}